\documentclass[11pt,a4paper]{article}

\usepackage{authblk}
\usepackage[hyperref]{emnlp2018}
\usepackage{times}
\usepackage{latexsym}
\usepackage{url}
\usepackage{framed}
\usepackage{todonotes}
\usepackage[normalem]{ulem}
\useunder{\uline}{\ul}{}
\usepackage[utf8]{inputenc}
\usepackage{booktabs}
\usepackage{multirow}

\aclfinalcopy

\author[1]{\textbf{James Thorne}}
\author[1]{\textbf{Andreas Vlachos}}
\affil[1]{Department of Computer Science, University of Sheffield}

\author[2]{\textbf{Oana Cocarascu}}
\affil[2]{Department of Computing, Imperial College London}

\author[3]{\\\textbf{Christos Christodoulopoulos}}
\author[3]{\textbf{Arpit Mittal}}
\affil[3]{Amazon Research Cambridge}

\affil[ ]{\texttt{\{j.thorne, a.vlachos\}@sheffield.ac.uk}}
\affil[ ]{\texttt{oana.cocarascu11@imperial.ac.uk}}
\affil[ ]{\texttt{\{chrchrs, mitarpit\}@amazon.co.uk}}

\title{The Fact Extraction and VERification (FEVER) Shared Task}
\begin{document}
\maketitle


\begin{abstract}
We present the results of the first Fact Extraction and VERification (FEVER) Shared Task. The task challenged participants to classify whether human-written factoid claims could be \textsc{Supported} or \textsc{Refuted} using evidence retrieved from Wikipedia. We received entries from 23 competing teams, 19 of which scored higher than the previously published baseline. The best performing system achieved a FEVER score of 64.21\%. In this paper, we present the results of the shared task and a summary of the systems, highlighting commonalities and innovations among participating systems.

\end{abstract}

\section{Introduction}
Information extraction is a well studied domain and the outputs of such systems enable many natural language technologies such as question answering and text summarization. However, 
since information sources can contain errors, there exists an additional need to verify whether the information is correct. 
For this purpose, we hosted the first Fact Extraction and VERification (FEVER) shared task to raise interest in and awareness of the task of automatic information verification - a research domain that is orthogonal to information extraction. This shared  task required participants to develop systems to predict the veracity of human-generated textual claims against textual evidence to be retrieved from Wikipedia. 

We constructed a purpose-built dataset for this task \cite{Thorne2018} that contains 185,445 human-generated claims, manually verified against the introductory sections of Wikipedia pages and labeled as \textsc{Supported}, \textsc{Refuted} or \textsc{NotEnoughInfo}. The claims were generated by paraphrasing facts from Wikipedia and mutating them in a variety of ways, some of which were meaning-altering. For each claim, and without the knowledge of where the claim was generated from, annotators selected evidence in the form of sentences from Wikipedia to justify the labeling of the claim. 

The systems participating in the FEVER shared task were required to label claims with the correct class and also return the 
sentence(s) forming the necessary evidence for the assigned label. Performing well at this task requires both identifying relevant evidence and reasoning correctly  with respect to the claim. A key difference between this task and other textual entailment and natural language inference tasks \cite{Dagan2009, Bowman2015} is the need to identify the evidence from a large textual corpus. Furthermore, in comparison to large-scale question answering tasks \cite{Chen2017}, systems must reason about information that is not present in the claim. We hope that research in these fields will be stimulated by the challenges present in FEVER. 

One of the limitations of using human annotators to identify correct evidence when constructing the dataset was the trade-off between annotation velocity and evidence recall \cite{Thorne2018}. Evidence selected by annotators was often incomplete. 
As part of the FEVER shared task, any evidence retrieved by participating systems that was not contained in the original dataset was annotated and used to augment the evidence in the test set. 

In this paper, we present a short description of the task and dataset, present a summary of the submissions and the leader board, 
and highlight future research directions.

\section{Task Description}
Candidate systems for the FEVER shared task were given a sentence of unknown veracity called a claim. The systems must identify suitable evidence from Wikipedia at the sentence level and assign a label whether, given the evidence, the claim is \textsc{Supported}, \textsc{Refuted} or whether there is \textsc{NotEnoughInfo} in Wikipedia to reach a conclusion. 
In 16.82\% of cases, claims required the combination of more than one sentence as supporting or refuting evidence. 
An example is provided in Figure~\ref{fig:ex}. 

\begin{figure}[t]
    \parskip=0pt
    \begin{framed}
    
    \parskip=0pt
      \begin{description}
\vspace{-0.1in}      
          \item[Claim:] The Rodney King riots took place in the most populous county in the USA. 
\vspace{-0.1in}          
          \item[\texttt{[wiki/Los\_Angeles\_Riots]}] The 1992 Los Angeles riots, \underline{also known as the Rodney King riots} were a series of riots, lootings, arsons, and civil disturbances that \underline{occurred in Los Angeles County}, California in April and May 1992.
\vspace{-0.1in}
\item[\texttt{[wiki/Los\_Angeles\_County]}] Los Angeles County, officially the County of Los Angeles, \underline{is the most populous county in the USA}. 
      \item[Verdict:] Supported

      \end{description}
      
    \parskip=0pt
\vspace{-0.2in}
\end{framed}

  \caption{Example claim from the FEVER shared task: a manually verified claim 
   that requires evidence from multiple Wikipedia pages.}
  \label{fig:ex}  
\end{figure}

\subsection{Data}
Training and development data was released through the FEVER website.\footnote{\url{http://fever.ai}} We used the reserved portion of the data presented in \citet{Thorne2018} as a blind test set. Disjoint training, development and test splits of the dataset were generated by splitting the dataset by the page used to generate the claim. The development and test datasets were balanced by randomly discarding claims from the more populous classes.

\begin{table}[h]
\centering
\begin{tabular}{@{}cccc@{}}
\toprule
\textbf{Split} & \textsc{Supported} & \textsc{Refuted} & \textsc{NEI} \\ \hline 
Training & 80,035 & 29,775 & 35,639 \\
Dev & 6,666 & 6,666 & 6,666 \\
Test & 6,666 & 6,666 & 6,666 \\
\bottomrule
\end{tabular}

\caption{Dataset split sizes for \textsc{Supported}, \textsc{Refuted} and \textsc{NotEnoughInfo} (\textsc{NEI}) classes}

\end{table}

\subsection{Scoring Metric}
We used the scoring metric described in \citet{Thorne2018} to evaluate the submissions. The FEVER shared task requires submission of evidence to justify the labeling of a claim. The training, development and test data splits contain multiple sets of evidence for each claim, each set being a minimal set of sentences that fully support or refute it. The primary scoring metric for the task is the label accuracy conditioned on providing at least one complete set of evidence, referred to as the FEVER score. Sentences labeled (correctly) as \textsc{NotEnoughInfo} do not require evidence. Correctly labeled claims with no or only partial evidence received no points for the FEVER score.  Where multiple sets of evidence was annotated in the data, only one set was required for the claim to be considered correct for the FEVER score.

Since the development and evaluation data splits are balanced, random baseline label accuracy ignoring the requirement for evidence is $33.33\%$. This performance level can also be achieved for the FEVER score by predicting \textsc{NotEnoughInfo} for every claim.
However, as the FEVER score requires evidence for \textsc{Supported} and \textsc{Refuted} claims, a random baseline is expected to score lower on this metric.

We provide an open-source release of the scoring software.\footnote{The scorer, test cases and examples can be found in the following GitHub repository \url{https://github.com/sheffieldnlp/fever-scorer}} Beyond the FEVER score, it computes precision, recall, $F_1$, and label accuracy to provide diagnostic information. The recall point is awarded, as is the case for the FEVER score, only by providing a complete set of evidence for the claim.

\subsection{Submissions}
The FEVER shared task was hosted as a competition on Codalab\footnote{\url{https://competitions.codalab.org/competitions/18814}} which allowed submissions to be scored against the blind test set without the need to publish the correct labels. The scoring system was open from 24th to 27th July 2018. Participants were limited to 10 submissions (max. 2 per day).\footnote{An extra half-day was given as an artifact of the competition closing at midnight pacific time.}

\begin{table*}[t]
\centering
\begin{tabular}{@{}clccccc@{}}
\toprule
\multirow{2}{*}{\textbf{Rank}} & \multirow{2}{*}{\textbf{Team Name}} & \multicolumn{3}{c}{\textbf{Evidence (\%)}}                                                   & \multirow{2}{*}{\textbf{\begin{tabular}[c]{@{}c@{}}Label \\ Accuracy (\%)\end{tabular}}} & \multirow{2}{*}{\textbf{\begin{tabular}[c]{@{}c@{}}FEVER\\ Score (\%)\end{tabular}}} \\
                               &                                     & \textbf{Precision}   & \multicolumn{1}{c}{\textbf{Recall}} & \multicolumn{1}{c}{\textbf{F1}} &                                                                                          &                                                                                      \\ \midrule

1             & UNC-NLP                                & 42.27                                                                      & 70.91                                                                   & 52.96                                                               & \textbf{68.21}                                                         & \textbf{64.21}                                                      \\
2             & UCL Machine Reading Group              & 22.16                                                                      & 82.84                                                          & 34.97                                                               & 67.62                                                                  & 62.52                                                               \\
3             & Athene UKP TU Darmstadt                & 23.61                                                                      & \textbf{85.19}                                                                   & 36.97                                                               & 65.46                                                                  & 61.58                                                               \\
4             & Papelo                                 & \textbf{92.18}                                                             & 50.02                                                                   & \textbf{64.85}                                                      & 61.08                                                                  & 57.36                                                               \\
5             & SWEEPer                                 & 18.48                                                                      & 75.39                                                                   & 29.69                                                               & 59.72                                                                  & 49.94                                                               \\
6             & Columbia NLP                           & 23.02                                                                      & 75.89                                                                   & 35.33                                                               & 57.45                                                                  & 49.06                                                               \\
7             & Ohio State University              & 77.23                                                                      & 47.12                                                                   & 58.53                                                               & 50.12                                                                  & 43.42                                                               \\
8             & GESIS Cologne                          & 12.09                                                                      & 51.69                                                                   & 19.60                                                               & 54.15                                                                  & 40.77                                                               \\
9             & FujiXerox                              & 11.37                                                                      & 29.99                                                                   & 16.49                                                               & 47.13                                                                  & 38.81                                                               \\
10            & \emph{withdrawn}                             & 46.60                                                                      & 51.94                                                                   & 49.12                                                               & 51.25                                                                  & 38.59                                                               \\
11            & Uni-DuE Student Team                                   & 50.65                                                                      & 36.02                                                                   & 42.10                                                               & 50.02                                                                  & 38.50                                                               \\
12            & Directed Acyclic Graph                 & 51.91                                                                      & 36.36                                                                   & 42.77                                                               & 51.36                                                                  & 38.33                                                               \\
13            & \emph{withdrawn}                                     & 12.90                                                                      & 54.58                                                                   & 20.87                                                               & 53.97                                                                  & 37.13                                                               \\
14            & Py.ro                                   & 21.15                                                                      & 49.38                                                                   & 29.62                                                               & 43.48                                                                  & 36.58                                                               \\
15            & SIRIUS-LTG-UIO                         & 19.19                                                                      & 70.82                                                                   & 30.19                                                               & 48.87                                                                  & 36.55                                                               \\
16            & \emph{withdrawn}                              & 0.00                                                                     & 0.01                                                                  & 0.00                                                              & 33.45                                                                  & 30.20                                                               \\
17            & BUPT-NLPer                               & 45.18                                                                      & 35.45                                                                   & 39.73                                                               & 45.37                                                                  & 29.22                                                               \\
18            & \emph{withdrawn}                                     & 23.75                                                                      & 86.07                                                                   & 37.22                                                               & 33.33                                                                  & 28.67                                                               \\
19            & \emph{withdrawn}                                  & 7.69                                                                       & 32.11                                                                   & 12.41                                                               & 50.80                                                                  & 28.40                                                               \\
20            & FEVER Baseline                         & 11.28                                                                      & 47.87                                                                   & 18.26                                                               & 48.84                                                                  & 27.45                                                               \\
21            & UMBC-FEVER                           & 49.01                                                                      & 29.66                                                                   & 36.95                                                               & 44.89                                                                  & 23.76                                                               \\
22            & \emph{withdrawn}                          & 26.81                                                                      & 12.08                                                                   & 16.65                                                               & 57.32                                                                  & 22.89                                                               \\
23            & \emph{withdrawn}                           & 26.33                                                                      & 12.20                                                                   & 16.68                                                               & 55.42                                                                  & 21.71                                                               \\
24            & University of Arizona                           & 11.28                                                                      & 47.87                                                                   & 18.26                                                               & 36.94                                                                  & 19.00                                                               \\ \bottomrule
\end{tabular}

\caption{Results on the test dataset.}
\label{tab:results}
\end{table*}

\section{Participants and Results}
86 submissions (excluding the baseline) were made to Codalab for scoring on the blind test set. There were 23 different teams which participated in the task (presented in Table~\ref{tab:results}).  19 of these teams scored higher than the baseline presented in \citet{Thorne2018}. All participating teams were invited to submit a description of their systems. We received 15 descriptions at the time of writing and the remaining are considered as \emph{withdrawn}. The system with the highest score was submitted by UNC-NLP (FEVER score: $64.21\%$). 

Most participants followed a similar pipeline structure to the baseline model. This consisted of three stages: document selection, sentence selection and natural language inference. However, some teams constructed models to jointly select sentences and perform inference in a single pipeline step, while others added an additional step, discarding inconsistent evidence after performing inference. 

Based on the team-submitted system description summaries (Appendix~\ref{sec:submitted}), in the following section we present an overview of which models and techniques were applied to the task and their relative performance. 

\section{Analysis}
\label{sec:analysis}

\subsection{Document Selection}
A large number of teams report a multi-step approach to document selection. The majority of submissions report extracting some combination of Named Entities, Noun Phrases and Capitalized Expressions from the claim. These were used either as inputs to a search API (i.e. Wikipedia Search or Google Search), search server (e.g.\ Lucene\footnote{\url{http://lucene.apache.org/}} or Solr\footnote{\url{http://lucene.apache.org/solr/}}) or as keywords for matching against Wikipedia page titles or article bodies. BUPT-NLPer report using \textsc{S-mart} for entity linking \citep{Yang2015} and the highest scoring team, UNC-NLP, report using page viewership statistics to rank the candidate pages. 
GESIS Cologne report directly selecting sentences using the Solr search, bypassing the need to perform document retrieval as a separate step. 

The team which scored highest on evidence precision and evidence F1 was Papelo (precision = $92.18\%$ and $F_1$ = $64.85\%$) who report using a combination of TF-IDF for document retrieval and string matching using named entities and capitalized expressions.


The 
teams which scored highest on evidence recall were Athene UKP TU Darmstadt (recall = $85.19\%$) and UCL Machine Reading Group (recall = $82.84\%$) \footnote{The withdrawn team that ranked 18th on $F_1$ score had the highest recall: $86.07\%$. A system description was not submitted by this team preventing us from including it in our analysis.} \footnote{The scores for precision, recall and $F_1$ were computed independent of the label accuracy and FEVER Score.} 
Athene report extracting noun-phrases from the claim and using these to query the Wikipedia search API. A similar approach was used by Columbia NLP who query the Wikipedia search API using named entities extracted from the claim as a query string, all the text before the first lowercase verb phrase as a query string and also combine this result with Wikipedia pages identified with Google search using the entire claim. UCL Machine Reading Group report a document retrieval approach that identifies Wikipedia article titles within the claim and ranks the results using features such as capitalization, sentence position and token match.

\subsection{Sentence Selection} There were three common approaches to sentence selection: keyword matching, supervised classification and sentence similarity scoring. Ohio State and UCL Machine Reading Group
report using keyword matching techniques: matching either named entities or tokens appearing in both the claim and article body. UNC-NLP, Athene UKP TU Darmstadt and Columbia NLP modeled the task as supervised binary classification, using architectures such as Enhanced LSTM \cite{Chen2016}, Decomposable Attention \cite{Parikh2016} or similar to them. SWEEPer and BUPT-NLPer present jointly learned models for sentence selection and natural language inference. Other teams report scoring based on sentence similarity using Word Mover's Distance \cite{Kusner2015} or cosine similarity over smooth inverse frequency weightings \cite{Arora2017}, ELMo embeddings \cite{Peters2018} and TF-IDF \cite{Salton1983}. 
UCL Machine Reading Group and Directed Acyclic Graph report an additional aggregation stage after the classification stage in the pipeline where evidence that is inconsistent is discarded.

\subsection{Natural Language Inference} 
NLI was modeled as supervised classification in all reported submissions. 
We compare and discuss the approaches for combining the evidence sentences together with the claim, sentence representations and training schemes. 
While many different approaches were used for sentence pair classification, e.g.\ Enhanced LSTM \cite{Chen2016}, Decomposable Attention \cite{Parikh2016}, Transformer Model \cite{Radford2018}, Random Forests \cite{svetnik2003random} and ensembles thereof, these are not specific to the task and it is difficult to assess their impact due to the differences in the processing preceding this stage. 

\paragraph{Evidence Combination:} UNC-NLP (the highest scoring team) concatenate the evidence sentences into a single string for classification; 
UCL Machine Reading Group classify each evidence-claim pair individually and aggregate the results using a simple multilayer perceptron (MLP); 
Columbia NLP perform majority voting;
and finally, Athene-UKP TU Darmstadt encode each evidence-claim pair individually using an Enhanced LSTM, pool the resulting vectors and use an MLP for classification. 

\paragraph{Sentence Representation:} University of Arizona explore using non-lexical features for predicting entailment, considering the proportion of negated verbs, presence of antonyms and noun overlap.  Columbia NLP learn universal sentence representations \citep{Conneau2017}. UNC-NLP include an additional token-level feature
the sentence similarity score from the sentence selection module. Both Ohio State and UNC-NLP report alternative token encodings: UNC-NLP report using ELMo \cite{Peters2018} and WordNet \cite{Miller1995} and Ohio State report using vector representations of named entities. FujiXerox report representing sentences using \textsc{DeIsTe} \cite{Yin2018}.

\paragraph{Training:} BUPT-NLPer and SWEEPer model the evidence selection and claim verification using a multi-task learning model under the hypothesis that information from each task supplements the other. SWEEPer also report parameter tuning using reinforcement learning.

\section{Additional Annotation}
As mentioned in the introduction, to increase the evidence coverage in the test set, the evidence submitted by participating systems was annotated by shared task volunteers after the competition ended. There were 18,846 claims where at least one system returned an incorrect label, according to the FEVER score, i.e.\ taking evidence into account.
These claims were sampled for annotation with a probability proportional to the number of systems which labeled each of them incorrectly. 

The evidence sentences returned by each system for each claim was sampled further with a probability proportional to the system's FEVER score in an attempt to focus annotation efforts towards higher quality candidate evidence.
These extra annotations were performed by volunteers from the teams participating in the shared task and three of the organizers.
Annotators were asked to label whether the retrieved evidence sentences supported or refuted the claim at question, and to highlight which sentences (if any), either individually or as a group, can be used as evidence. We retained the annotation guidelines from \citet{Thorne2018} (see Sections A.7.1, A.7.3 and A.8 from that paper for more details).


At the time of writing, 1,003 annotations were collected for 618 claims. This identified 3 claims that were incorrectly labeled as \textsc{Supported} or \textsc{Refuted} and 87 claims that were originally labeled as \textsc{NotEnoughInfo} that should be re-labeled as \textsc{Supported} or \textsc{Refuted} through the introduction of new evidence (44 and 43 claims respectively). 308 new evidence sets were identified for claims originally labeled as \textsc{Supported} or \textsc{Refuted}, consisting of 280 single sentences and 28 sets of 2 or more sentences.

Further annotation is in progress and the data collected as well as the final results will be made public at the workshop.


\section{Conclusions}
The first Fact Extraction and VERification shared task attracted submissions from 86 submissions from 23 teams. 19 of these teams exceeded the score of the baseline presented in \citet{Thorne2018}. For the teams which provided a system description, we highlighted the approaches, identifying commonalities and features that could be further explored. 

Future work will address limitations in human-annotated evidence and explore other subtasks needed to predict the veracity of information extracted from untrusted sources.

\section*{Acknowledgements}
The work reported was partly conducted while James Thorne was at Amazon Research Cambridge. Andreas Vlachos is supported by the EU H2020 SUMMA project (grant agreement number 688139).

\bibliography{references}
\bibliographystyle{acl_natbib_nourl}

\appendix
\section{Short System Descriptions Submitted by Participants}
\label{sec:submitted}

\subsection{UNC-NLP}
Our system is composed of three connected components namely, a document retriever, a sentence selector, and a claim verifier. The document retriever chooses candidate wiki-documents via matching of keywords between the claims and the wiki-document titles, also using external page-view frequency statistics for wiki-page ranking. The sentence selector is a sequence-matching neural network that conducts further fine-grained selection of evidential sentences by comparing the given claim with all the sentences in the candidate documents. This module is trained as a binary classifier given the ground truth evidence as positive examples and all the other sentences as negative examples with an annealing sampling strategy. Finally, the claim verifier is a state-of-the-art 3-way neural natural language inference (NLI) classifier (with WordNet and ELMo features) that takes the concatenation of all selected evidence as the premise and the claim as the hypothesis, and labels each such evidences-claim pair as one of `support', `refute', or `not enough info'. To improve the claim verifier via better awareness of the selected evidence, we further combine the last two modules by feeding the sentence similarity score (produced by the sentence selector) as an additional token-level feature to the claim verifier.

\subsection{UCL Machine Reading Group}

The UCLMR system is a four stage model consisting of document retrieval, sentence retrieval, natural language inference and aggregation. Document retrieval attempts to find the name of a Wikipedia article in the claim, and then ranks each article based on capitalization, sentence position and token match features. A set of sentences are then retrieved from the top ranked articles, based on token matches with the claim and position in the article. A natural language inference model is then applied to each of these sentences paired with the claim, giving a prediction for each potential evidence. These predictions are then aggregated using a simple MLP, and the sentences are reranked to keep only the evidence consistent with the final prediction.

\subsection{Athene UKP TU Darmstadt}

\paragraph{Document retrieval}
We applied the constituency parser from AllenNLP to extract noun phrases in the claim and made use of Wikipedia API to search corresponding pages for each noun phrase. So as to remove noisy pages from the results, we have stemmed the words of their titles and the claim, and then discarded pages whose stemmed words of the title are not completely included in the set of stemmed words in the claim. 

\paragraph{Sentence selection}
The hinge loss with negative sampling is applied to train the enhanced LSTM. For a given positive claim-evidence pair, negative samples are generated by randomly sampling sentences from the retrieved documents. 

\paragraph{RTE} 
We combine the 5 sentences from sentence selection and the claim to form 5 pairs and then apply enhanced LSTM for each pair. 
We combine the resulting representations using average and max pooling and feed the resulting vector through an MLP for classification.

\subsection{Papelo}
We develop a system for the FEVER fact extraction and verification
challenge that uses a high precision entailment classifier based on
transformer networks pretrained with language modeling \cite{Radford2018}, to classify a
broad set of potential evidence.  The precision of the entailment classifier
allows us to enhance recall by considering every statement from several
articles to decide upon each claim.  We include not only the articles best
matching the claim text by TFIDF score, but read additional articles whose
titles match named entities and capitalized expressions occurring in the
claim text.  The entailment module evaluates potential evidence one statement
at a time, together with the title of the page the evidence came from
(providing a hint about possible pronoun antecedents).  In preliminary
evaluation, the system achieved 57.36\% FEVER score, 61.08\% label accuracy,
and 64.85\% evidence F1 on the FEVER shared task test set.

\subsection{SWEEPer}
\label{team:ucl}
Our model for fact checking and verification consists of two stages: 1) identifying relevant documents using lexical and syntactic features from the claim and first two sentences in the Wikipedia article and 2) jointly modeling sentence extraction and verification.  As the tasks of fact checking and finding evidence are dependent on each other, an ideal model would consider the veracity of the claim when finding evidence and also find only the evidence that supports/refutes the position of the claim.  We thus jointly model the second stage by using a pointer network with the claim and evidence sentence represented using the ESIM module.  For stage 2, we first train both components using multi-task learning over a larger memory of extracted sentences, then tune parameters using reinforcement learning to first extract sentences and predict the relation over only the extracted sentences.

\subsection{Columbia NLP}

For document retrieval we use three components: 1) use google custom search API with the claim as a query and return the top 2 Wikipedia pages; 2) extract all name entities from the claims and use Wikipedia python API to return a page for each name entity and 3); use the prefix of the claim until the first lowercase verb phrase, and use Wikipedia API to return the top page. 

For Sentence Selection we used the modified document retrieval component of DrQA to get the top 5 sentences and then further extracted the top 3  sentences using cosine similarity between vectors obtained from Elmo \cite{Peters2018}  sentence embeddings of the claim and the evidence. 

For RTE we used the same model as outlined by \cite{Conneau2017} in their work for recognizing textual entailment and learning universal sentence representations. If at least one out of the three evidences SUPPORTS/REFUTES the claim and the rest are NOT ENOUGH INFO , then we treat the label as SUPPORTS/REFUTES, else we return the majority among three classes as the predicted label.

\subsection{Ohio State University}

Our system was developed using a heuristics-based approach for evidence extraction and a modified version of the inference model by \citet{Parikh2016} for classification into refute, support, or not enough info. Our process is broken down into three distinct phases. First, potentially relevant documents are gathered based on key words/phrases in the claim that appear in the wiki dump. Second, any possible evidence sentences inside those documents are extracted by breaking down the claim into named entities plus nouns and finding any sentences which match those entities, while allowing for various exceptions and additional potential criteria to increase recall. Finally, every sentences is classified into one of the three categories by the inference tool, after additional vectors are added based on named entity types. NEI sentences are discarded and the highest scored label of the remaining sentences is assigned to the claim.

\subsection{GESIS Cologne}
In our approach we used a sentence wise approach in all components.
To find the sentences we set up a Solr database and indexed every sentence including information about the article where the sentence is from.
We created queries based on the named entities and noun chunks of the claims.
For the entailment task we used a Decomposable Attention Model similar to the one used in the baseline approach.
But instead of comparing the claim with all top 5 sentences at once we treat every sentence separately.
The results for the top 5 sentence where then joined with an ensemble learner incl. the rank of the sentence retriever of the wikipedia
sentences.

\subsection{FujiXerox}
We prepared a pipeline system which composes of document selection, a sentence retrieval, and a recognizing textual entailment (RTE) components. A simple entity linking approach with text match is used as the document selection component, this component identifies relevant documents for a given claim by using mentioned entities as clues. The sentence retrieval component selects relevant sentences as candidate evidence from the documents based on TF-IDF. Finally, the RTE component selects evidence sentences by ranking the sentences and classifies the claim as SUPPORTED, REFUTED, or NOTENOUGHINFO simultaneously. As the RTE component, we adopted DEISTE (Deep Explorations of Inter-Sentence interactions for Textual Entailment) \cite{Yin2018} model that is the state-of-the-art in RTE task.

\subsection{Uni-DuE Student Team}
We generate a Lucene index from the provided Wikipedia dump. 
Then we use two neural networks, one for named entity recognition and 
the other for constituency parsing, and also the Stanford dependency 
parser to create the keywords used inside the Lucene queries. 
Depending on the amount of keywords found for each claim, we run 
multiple Lucene searches on the generated index to create a list of candidate sentences for each claim. 
The resulting list of claim-candidate pairs is processed in three ways: 
\begin{enumerate}
\item We use the Standford POS-Tagger to generate POS-Tags for the claim 
and candidate sentences which are then used in a handcrafted scoring 
script to assign a score on a 0 to 15 scale. 
\item We run each pair through a modified version of the Decomposable 
Attention network. 
\item We merge all candidate sentences per claim into one long piece of 
text and run the result paired with the 
claim through the same modified Decomposable Attention network as in 
(2.). 
\end{enumerate}

We then make the final prediction in a handcrafted script combining the 
results of the three previous steps.

\subsection{Directed Acyclic Graph}

In this paper, we describe the system we designed for the FEVER 2018 Shared Task. The aim of this task was to conceive a system that can not only automatically assess the veracity of a claim but also retrieve evidence supporting this assessment from Wikipedia. 
In our approach, the Wikipedia documents whose Term Frequency - Inverse Document Frequency (TFIDF) vectors are most similar to the vector of the claim and those documents whose names are similar to the named entities (NEs) mentioned in the claim are identified as the documents which might contain evidence. 
The sentences in these documents are then supplied to a decomposable attention-based textual entailment recognition module. This module calculates the probability of each sentence supporting the claim, contradicting the claim or not providing any relevant information. Various features computed using these probabilities are finally used by a Random Forest classifier to determine the overall truthfulness of the claim. The sentences which support this classification are returned as evidence. Our approach achieved a 42.77\% evidence F1-score, a 51.36\% label accuracy and a 38.33\% FEVER score.

\subsection{Py.ro}
We NER tagged the claim using SpaCy and used the Named Entities as candidate page IDs. We resolved redirects by following the Wikipedia URL if an item was not in the preprocessed dump. If a page could not be found, we fell back to the baseline document selection method. The rest of the system was identical to the baseline system, although we used our document retrieval system to generate alternative training data.

\subsection{SIRIUS-LTG-UIO}

This  article  presents  the  SIRIUS-LTG  system  for  the  Fact  Extraction  and  VERification (FEVER)  Shared  Task.   Our  system  consists of three components: 
\begin{enumerate}
\item Wikipedia Page Retrieval:   First  we  extract  the  entities  in  the claim,  then we find potential Wikipedia URI candidates  for  each  of  the  entities  using  the SPARQL  query  over  DBpedia  
\item Sentence selection:   We  investigate  various  techniques i.e.   SIF embedding,  Word Mover’s Distance (WMD),  Soft-Cosine  Similarity,  Cosine similarity with unigram TF-IDF to  rank  sentences  by  their  similarity  to  the  claim. 
\item Textual Entailment:  We  compare  three  models  for  the  claim  classification.   We  apply  a Decomposable Attention (DA) model \cite{Parikh2016},  a  Decomposed  Graph  Entailment (DGE) model \cite{Khot2018} and a Gradient-Boosted Decision Trees (TalosTree) model  \cite{Baird2017}  for  this  task.   
\end{enumerate}
The experiments show  that the pipeline with simple Cosine Similarity using TFIDF in sentence selection along  with  DA  as  labeling model achieves better results in development and test dataset.

\subsection{BUPT-NLPer}
We introduce an end-to-end multi-task learning model for fact extraction and verification with bi-direction attention. We propose a multi-task learning framework for the evidence extraction and claim verification because these two tasks can be accomplished at the same time. Each task provides supplementary information for the other and improves the results of another task.

For each claim, our system firstly uses the entity linking tool S-MART to retrieve relative pages from Wikipedia. Then, we use attention mechanisms in both directions, claim-to-page and page-to-claim, which provide complementary information to each other. Aimed at the different task, our system obtains claim-aware sentence representation for evidence extraction and page-aware claim representation for claim verification.

\subsection{University of Arizona}

Many approaches to automatically recognizing entailment relations have employed classifiers over hand engineered lexicalized features, or deep learning models that implicitly capture lexicalization through word embeddings. This reliance on lexicalization may complicate the adaptation of these tools between domains. For example, such a system trained in the news domain may learn that a sentence like ``Palestinians recognize Texas as part of Mexico'' tends to be unsupported, a fact which has no value in say a scientific domain. To mitigate this dependence on lexicalized information, in this paper we propose a model that reads two sentences, from any given domain, to determine entailment without using any lexicalized features. Instead our model relies on features like proportion of negated verbs, antonyms, noun overlap etc. In its current implementation, this model does not perform well on the FEVER dataset, due to two reasons. First, for the information retrieval part of the task we used the baseline system provided, since this was not the aim of our project. Second, this is work in progress and we still are in the process of identifying more features and gradually increasing the accuracy of our model. In the end, we hope to build a generic end-to-end classifier, which can be used in a domain outside the one in which it was trained, with no or minimal re-training.

\subsection{UMBC-FEVER}
We describe the UMBC-FEVER system that we used in the 2018 FEVER shared task. The system employed a frame-based information retrieval approach
to select Wikipedia sentences providing evidence and used a two-layer multilayer perceptron (MLP) for classification. Our submission achieved a score 
of 0.3695 on the Evidence F1 metric for retrieving relevant evidential sentences (10$^{th}$ out of 24) and a score of 0.2376 on the FEVER metric 
(just below the baseline system).


\end{document}